%% file: lrec2018.tex
\documentclass[10pt, a4paper]{article}
\usepackage{lrec}
\usepackage{multibib}
\newcites{languageresource}{Language Resources}
\usepackage{graphicx}
\usepackage{tabularx}
\usepackage{soul}

\usepackage{epstopdf}
\usepackage[latin1]{inputenc}

\usepackage{hyperref}
\usepackage{xstring}
\usepackage{caption}
\usepackage{algorithmicx}
\usepackage[ruled]{algorithm}
\usepackage{algpseudocode}
\usepackage{multicol}
\usepackage{multirow}
\usepackage{booktabs}
\usepackage{stfloats}
\usepackage{color}
\usepackage[export]{adjustbox}

\hypersetup{
  pdftitle = {Live Blog Corpus for Summarization},
  pdfauthor = {Avinesh P.V.S., Maxime Peyrard, Christian M. Meyer},
  pdfsubject  = {The 11th edition of the Language Resources and Evaluation Conference (LREC)},
  pdfkeywords = {Natural Language Processing, Language Resources, Live blogs, Summarization Corpus, Corpus Construction, Focused Crawling, Online Journalism}
}

\title{Live Blog Corpus for Summarization}

\name{Avinesh P.V.S., Maxime Peyrard, Christian M. Meyer}

\address{Research Training Group AIPHES and UKP Lab \\
Computer Science Department, Technische Universit{\"a}t Darmstadt \\
www.aiphes.tu-darmstadt.de, www.ukp.tu-darmstadt.de}
         
\abstract{
Live blogs are an increasingly popular news format to cover breaking news and live events in online journalism. 
Online news websites around the world are using this medium to give their readers a minute by minute update on an event.
Good summaries enhance the value of the live blogs for a reader but are often not available.
In this paper, we study a way of collecting corpora for automatic live blog summarization.
In an empirical evaluation using well-known state-of-the-art summarization systems, we show that live blogs corpus poses new challenges in the field of summarization.
We make our tools publicly available to reconstruct the corpus to encourage the research community and replicate our results. 
\url{https://github.com/UKPLab/lrec2018-live-blog-corpus}
\\ \newline  }

\begin{document}

\maketitleabstract

\section{Introduction}
\label{intro}
\input{introduction}

\section{Related Work}
\label{sec:related}
\input{related_work}

\section{Corpus Construction}
\label{sec:approach}
\input{approach}

\section{Corpus Statistics}
\label{sec:corpus_statistics}
\input{corpus_details}

\section{Results and Analysis}
\label{sec:setup}
\input{experiments}

\section{Conclusion and Future Work}
\label{sec:conclusion}
\input{conclusion}

\section*{Acknowledgments}
This work has been supported by the German Research Foundation as part of the Research Training Group ``Adaptive Preparation of Information from Heterogeneous Sources'' (AIPHES) under grant No. GRK 1994/1. We also acknowledge the useful comments and suggestions of the anonymous reviewers.

\section{Bibliographical References}
\label{main:ref}

\bibliographystyle{lrec}
\bibliography{lrec2018}


\end{document}

%% file: introduction.tex
\begin{figure}[h]
\includegraphics[width=0.5\textwidth]{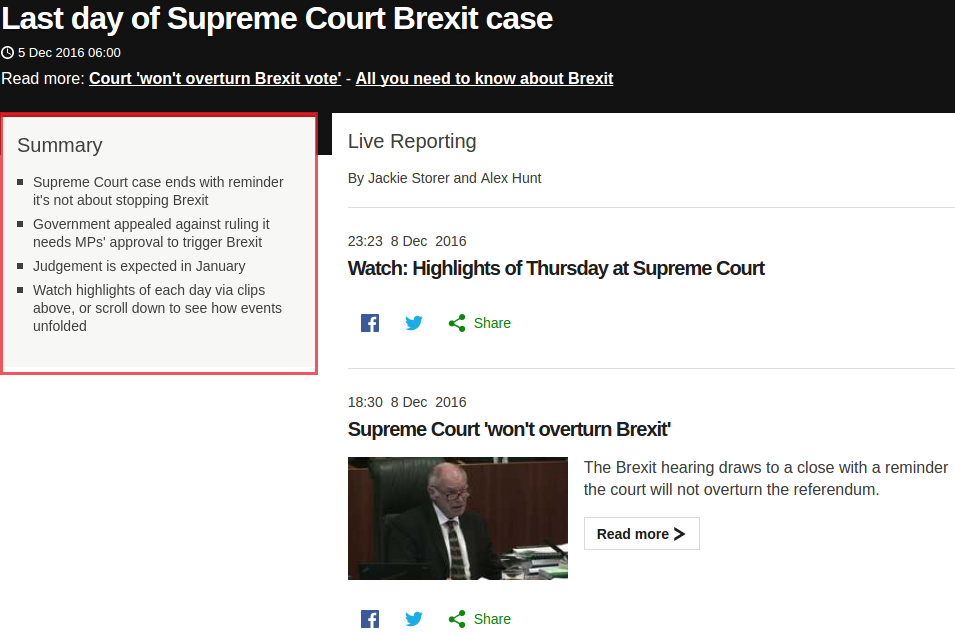}
\caption{BBC.com live blog on ``Last day of Supreme Court Brexit Case"}
\label{fig:blog_examples1}
\end{figure}
\begin{figure}[t]
\includegraphics[width=0.5\textwidth]{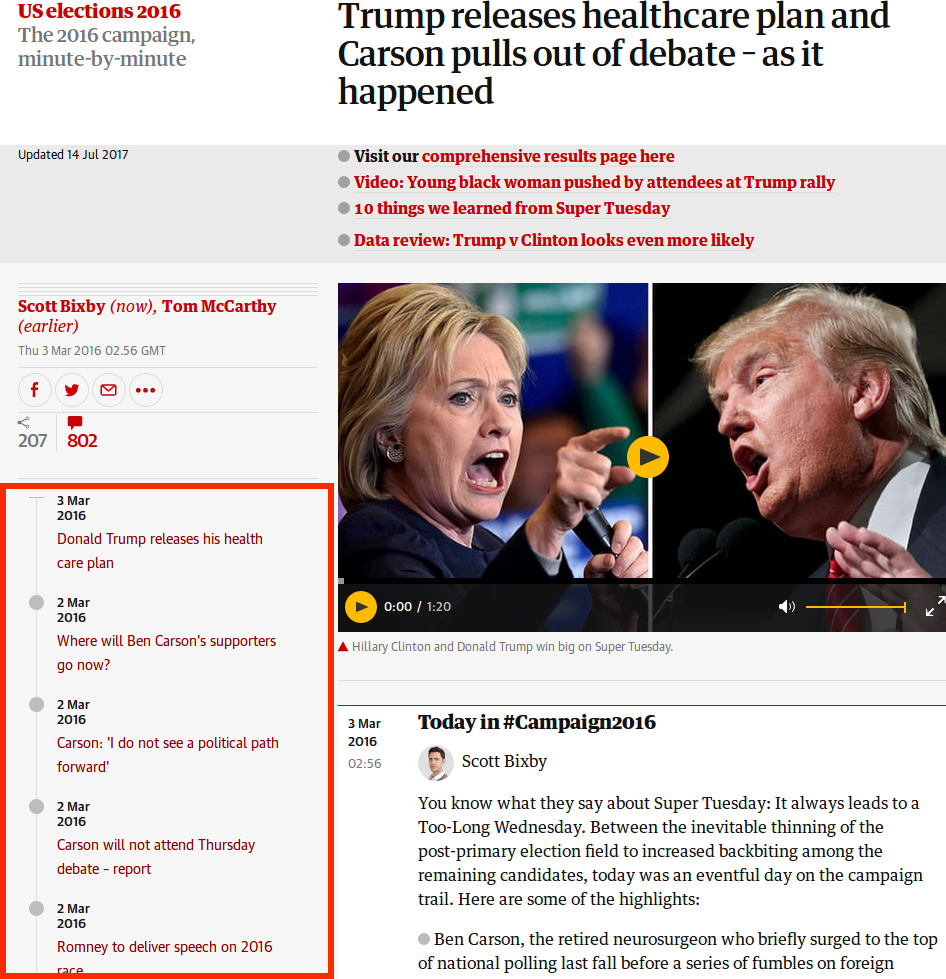}
\caption{TheGuardian.com live blog on ``US elections 2016 campaign''}
\label{fig:blog_examples2}
\end{figure}

A live blog is a dynamic news article providing a rolling textual coverage of an ongoing event. 
It is a single article continuously updated by one or many journalists with timestamped micro-updates typically displayed in chronological order.
Live blogs can contain a wide variety of media, including text, video, audio, images, social media snippets and links.
At the end of the broadcasting, a journalist usually summarizes the main information about the event. 
For more extended events, journalists may also write intermediate summaries.
Figure \ref{fig:blog_examples1} and \ref{fig:blog_examples2} show an example live blog provided by the \emph{BBC} on ``Last day of Supreme Court Brexit Case'' and \emph{The Guardian} on ``US elections 2016 campaign''. The timestamped information snippets are on the right, the human-written bullet-point summary is at the top left.

In the last decade, live-blogging has become very popular. 
It is commonly used by major news organizations, such as the \emph{BBC}, \emph{The Guardian} or \emph{The New York Times}. 
Several different kinds of events are regularly covered by live blogs, including sport games, elections, ceremonies, protests, conflicts and natural disasters. 
\newcite[p.1]{Thurman2016} report a journalist's view that ``live blogs have transformed the way we think about news, our sourcing, and everything''.
Thanks to this new journalistic trend, many live blogs -- and their human-written summaries -- are available online and new ones are generated every day. 

In this work, we propose to leverage this data and investigate the task of automatic live blog summarization by crawling a new dataset. 
Live blog summarization has more direct applications in Journalism than the traditional but rather artificial tasks of a single document and multi-document summarization. 
Systems capable of summarizing live streams of heterogeneous content can be directly beneficial to users and even assist journalists during their daily work.

However, this new task also comes with new challenges. 
Live blogs are a list of short snippets of heterogeneous information and they do not form one coherent piece of text. 
The non-cohesive snippets make the task different from single document summarization. 
Furthermore, most single documents are easily summarized by the baseline extracting the first few sentences. 
Such an approach is not effective for live blogs due to their heterogeneity and chronological order. 
The snippets are typically small, focused, numerous and rarely redundant which contrasts with the well-studied task of multi-document summarization. 
The topic is continually shifting and many sub-topics may arise and become central at some point. This even differs from the single topic shift found in classical update summarization tasks.
For example, the live blog in Figure \ref{fig:blog_examples1} on ``Last day of Supreme Court Brexit case'' consists of topic shift across ``Supreme court judgment'', ``Government appeal'', ``Opinions of MP's on Brexit'' and others.
Moreover, when summarizing a live blog, one has to account for the whole past and all sub-topics previously discussed, which differs from real-time summarization setups like TREC. 


We focus on two online news websites for acquiring live blogs, the \emph{BBC}\footnote{\url{http://www.bbc.com}} and \emph{The Guardian}\footnote{\url{https://www.theguardian.com}}, because they contain a lot of easily accessible live blogs that we automatically crawl and process.

In summary, our contributions are:
\begin{itemize}
\item We introduce a new task: live blog summarization.
\item We suggest a pipeline to collect and extract live blogs with human-written summaries from two major online newspapers and release it for the community\footnote{\url{https://github.com/UKPLab/lrec2018-live-blog-corpus}}.
\item We benchmark the dataset with commonly used summarization methods to stimulate further research into this challenging task.
\end{itemize}

The rest of the paper is structured as follows: Section \ref{sec:related} details existing summarization corpora and related works.
Section~\ref{sec:approach} discusses our approach to collect live blogs from \emph{BBC} and \emph{The Guardian}, followed by a discussion on the statistics and properties of our live blog corpus in section \ref{sec:corpus_statistics}.
The performance of well-established summarization baselines on this new dataset is discussed in section \ref{sec:setup}, followed by the conclusion and future work.

%% file: related_work.tex
In this section, we describe previous works related to summarization corpora. 
They were focused on single and multi-document summarization, update summarization, and real-time summarization.
We are not aware of any previous work on live blog summarization.

\paragraph{Single and multi-document summarization.}
The most widely used summarization datasets have been published in the Document Understanding Conference\footnote{\url{http://duc.nist.gov/}} (DUC) series. 
In total, there are 139 document clusters with 376 human-written reference summaries across DUC '01, '02, and '04.
Although the research community has often used these corpora, creating the manual summaries is time-consuming and labor-intensive.

Large datasets typically exist for single document summarization tasks, for example, the ACL Anthology Reference Corpus \cite{Bird08} and the CNN/Daily Mail dataset \cite{Hermann2015}.
The latter contains large pairs of 312k online news articles and multi-sentence summaries used for neural summarization approaches \cite{Nallapati2016,See2017}.
However, their dataset contains only one source document, whereas live blogs have a larger number of information snippets, typically more than 100.

Another recent work uses social media's reactions on Twitter to create large-scale multi-document summaries for news \cite{Lloret2013,Cao2016}.
\newcite{Cao2016} use hashtags to cluster the documents into the same topic and use tweets with hyperlinks to generate optimal reference summaries.
Their corpus consists of 204 document clusters with 1,114 documents and 4,658 reference tweets. 
Although this approach uses social media information to create a summarization corpus, they produce synthetic summaries, which are not written by a human. 
Moreover, they only use the corpus for training supervised learning approaches and not for evaluating summarization systems. 

Other multi-document summarization datasets focus on heterogeneous sources \cite{Zopf2016,Benikova16,Nakano10}, multiple languages \cite{Giannakopoulos15}, and reader-aware multi-document summaries \cite{Li2017}, which jointly aggregate news documents and reader comments.  

\paragraph{Update summarization.}
After the DUC series, the Text Analysis Conference\footnote{\url{http://www.nist.gov/tac/}} (TAC) series ('08, '09) introduced the update summarization task \cite{Dang2008}. 
In this task, two summaries are provided for two sets of documents and the summary of the second set of documents is an update of the first set.
Although the importance of text to be included in the summary solely depends on the novelty of the information, the task usually observes only a single topic shift. 
In live blogs, however, there are multiple sub-topics and the importance of the sub-topics changes over time.

\paragraph{Real-time summarization.}
Real-time summarization began at the Text REtrieval Conference\footnote{\url{http://trec.nist.gov/}} (TREC) 2016 and represents an amalgam of the microblog track and the temporal summarization track \cite{Lin2016}.
In real-time summarization, the goal is to automatically monitor the stream of documents to keep a user up to date on topics of interest and create email digests that summarize the events of that day for their interest profile. 
The drawback of this task is that they have a predefined time frame for evaluation due to the real-time constraint, which makes the development of systems and replicating results arduous.
Note that live blog summarization is very similar to real-time summarization, as the real-time constraint also holds true for live blogs if the summarization system is applied to the stream of snippets.
Moreover, the Guardian live blogs do consist of updated and real-time summaries, but this requires different real-time crawling strategies which are out of the scope of this work.

%% file: approach.tex
In this section, we describe the three steps to construct our live blogs summarization corpus:
(1) live blog crawling yielding a list of URLs,
(2) content parsing and processing, where the documents and corresponding summaries with the metadata are extracted from the URLs and stored in a JSON format, and 
(3) live blog pruning as a final step for creating a high-quality gold standard live blog summarization corpus.

\paragraph{Live blog Crawling.}
On the Guardian, a frequently updated index webpage\footnote{\url{http://www.theguardian.com/tone/minutebyminute}} references all archived live blogs.
We took a snapshot of this page that provided us with 16,246 unique live blogs.

In contrast, the BBC website has no such live blog archive.
Thus, we use an iterative approach similar to BootCaT \cite{Baroni2004} as described in Algorithm~\ref{LiveBlogRetrieval} to bootstrap a corpus utilizing a set of seed terms extracted from ten BBC live blog links from the web.
The iterative procedure starts with a small set of seed terms ($K_0$) and gathers new live blog links using automated Bing queries\footnote{\url{https://azure.microsoft.com/en-us/services/cognitive-services/bing-web-search-api}} by exploiting patterns ($P$) in live blog URLs (i.e. ``site:http://www.bbc.com/news/live/[key term]'' as in line \ref{line:make_queries}). 
We collect all the valid links returned by the Bing queries (line \ref{line:search}) and look for new key terms in the recently retrieved live blogs (line \ref{line:extract_terms}).
In our implementation, key terms are terms with high TF*IDF scores.
The new key terms are then used in the Bing queries of the subsequent iterations.
The process is repeated until no new live blogs are discovered anymore (line \ref{line:stop_criteria}).
With this process, we ran 4,000 search queries returning each around 1,000 results on average and we collected 9,931 unique URLs.

Although our method collected a majority of the live blogs in the 4,000 search queries, a more sophisticated key terms selection could minimize the search queries and maximize the unique URLs.
Additionally, this methodology can be applied to other news websites featuring live blogs like \emph{The New York Times}, \emph{Washington Post} or \emph{Der Spiegel}.

An important point to note is that we find the collected BBC live blog URLs predominantly cover more recent years. 
This usage could be due to the Bing Search API preferring recent articles for the first 100 results.
To collect a broad range of news articles the queries need to be precise.


\begin{algorithm}
\caption{Iterative Live Blog Retrieval}\label{LiveBlogRetrieval}
\begin{algorithmic}[1]
\algrenewcommand\algorithmicindent{1.25em}
\Procedure{LiveBlogRetrieval()}{}
\State \textbf{input:} \textrm{Seed terms} $K_0$, \textrm{Live blog Pattern} $P$
\State $L_0 \gets \emptyset$
\For{$t = 1...T$} \label{line:loop}
\State $Q_t \gets \textrm{makeQueries}(K_{t-1}, P)$ \label{line:make_queries}
\State $L_t \gets \textrm{getLinks}(Q_t)$ \label{line:search}
\If {$ \cup_{i=0}^{t-1}L_i = \cup_{i=0}^{t}L_i $} \label{line:stop_criteria}
  \State \textbf{return} $\cup_{i=0}^{t}L_i$
\Else
  \State $K_{t} \gets \textrm{extractKeyTerms}(L_t) - \cup_{i=0}^{t-1}K_i$ \label{line:extract_terms}
\EndIf
\EndFor
\EndProcedure
\end{algorithmic}
\end{algorithm}

\paragraph{Content Parsing and Processing.} 
Once the URLs are retrieved, we fetch the HTML content, remove the boiler-plate and store the cleaned data in a JSON file.

During this step, unreachable URLs were filtered out.
We discard live blogs for which we could not retrieve the summary or correctly parse the information snippets.
Indeed, live blogs can have changing patterns over time rendering the automatic extraction difficult. 

Parsing of BBC live blogs can be automated easily because both bullet-point summaries and information snippets follow a consistent pattern. 
For the Guardian, we identify several recurring patterns which cover most of the live blogs.
The Guardian live blogs were in use since 2001 but were in experimental phase till 2008.
Due to the lack of a specific structure or a summary during this experimental phase, we remove 10k of the crawled live blogs. 
However, after 2008, live blogs have had a prominent place in the editorial with a consistent structure.

We parse metadata like URL, author, date, genre, summaries and documents for each live blog using site-specific regular expressions on the HTML source files.

After this step, 7,307 live blogs remain for BBC and 6,450 for Guardian.

\paragraph{Live blog Pruning.}
To further clean the data, we decided to remove live blogs exhibiting several topics as they can be quite noisy. 
For example, BBC provides some live blogs covering all events happening in a given region within a given time frame (e.g., \emph{Essex: Latest updates}). 
We also prune live blogs about sport games and live chats, because the summaries are based on simple templates.

We further prune live blogs based on their summaries. 
We first remove a sentence of a summary if it has less than three words. 
Then, we discarded live blogs whose summaries have less than three sentences. 
This is to ensure the quality of the corpus, as overly short summaries would yield a different summarization goal similar to headline generation and they are typically an indicator for a non-standard live blog layout.

After the whole pruning step, 974 live blogs remained for BBC and 1,681 for the Guardian.

Overall, 10\% of the initial set of live blogs, both for BBC and Guardian remained after selective pruning.
This is to ensure high-quality summaries for the live blogs.
Although the pruning rejects 90\% of the live blogs, the size of the live blog corpus is 20--30 times larger than the classical corpora released during DUC, TREC and TAC tasks. 

\begin{table}
\centering
{
\begin{tabular}{lrrr}
\toprule
Dataset & Crawling & Processing & Pruning\\
\midrule
BBC    	  & 9,931  & 7,307  & 974 \\
Guardian  & 16,246 & 6,405 & 1,681\\
\bottomrule
\end{tabular}}
\caption{Number of topics for BBC and the Guardian}
\label{tab:stats}
\end{table}

\paragraph{Code Repository.}

To replicate our results and advance research in live blog summarization we publish our tools for reconstructing the live blog corpus open-source under the Apache License 2.0.
The repository consists of 
(a) raw and processed URLs, 
(b) tools for crawling live blogs,
(c) tools for parsing the content of the URLs and transforming content into JSON, and
(d) code for calculating baselines and corpus statistics.

%% file: corpus_details.tex
We compute several statistics about the corpora and report them in Table \ref{tab:counts}.
The number of documents (or snippets) per topic is around 95 for BBC and 56 for the Guardian.
In comparison, standard multi-document summarization datasets like DUC '04\footnote{\url{http://duc.nist.gov/duc2004}} and TAC '08A\footnote{\url{https://tac.nist.gov/2008}} have only 10 documents per topic. 

Furthermore, we observe that snippets are quite short as there is an average of 62 words per snippet for BBC and 108 for the Guardian.
Summaries are also shorter than summaries in standard datasets. Indeed, in DUC2004 and TAC2008A summaries are expected to contain 100 words.

Our corpora are larger because, together, they contain 2,655 topics and 186,999 documents. With many data points, machine learning approaches become readily applicable. 

\begin{table}[!ht]
\centering
{
\begin{tabular}{lrr}
\toprule
Statistic & BBC & Guardian \\
\midrule
\# topics    	  	& 974   & 1,681 \\
\# documents       	& 92,537  & 94,462 \\
\# documents / topic 	& 95.01  & 56.19 \\
\# words / document 	& 61.75  & 107.53 \\
\# words / summary 	& 59.48  & 42.23 \\
\bottomrule
\end{tabular}}
\caption{Corpus statistics for BBC and the Guardian}
\label{tab:counts}
\end{table}

\paragraph{Domain Distribution.}

Live blogs cover a wide range of subjects from multiple domains. 
In Table \ref{tab:domain}, we report the distribution of different domains in our combined datasets (BBC and the Guardian). 
While we observe that politics, business and news are the most prominent domains, there is also a number of well-represented domains like local and international events or culture.

\begin{table}[!t]
\centering
{
\begin{tabular}{lrr}
\toprule
Domain & \# topics & proportion (\%) \\
\midrule
Politics    	  	& 834   & 31.41 \\
Business 		& 421 	& 15.86 \\
General News            & 369  &  13.90 \\
UK local events		& 368  & 13.86 \\
International events	& 337  & 12.69 \\
Culture 		& 186  & 7.01 \\
Science 		& 60  & 2.26 \\
Society 		& 27  & 1.02 \\
Others 			& 53  & 2.00 \\
\bottomrule
\end{tabular}}
\caption{Corpus distribution across multiple domains for BBC and the Guardian}
\label{tab:domain}
\end{table}

\paragraph{Heterogeneity.}

The resulting corpus is expected of exhibiting various levels of heterogeneity.
Indeed, there contain various topics with mixed writing styles (short-to-the-point snippets vs.\ longer descriptive snippets).
Furthermore, live blogs are subject to topic shifts which could be observed by the change in words used.

To measure this textual heterogeneity, we use information theoretic metrics on word probability distributions like it was done before in analyzing the heterogeneity of summarization corpora \cite{Zopf2016}.
Based on Jensen-Shannon (JS) divergence, they defined a measure of textual heterogeneity $TH$ for a topic $T$ composed of documents $d_1, \cdots, d_n$ as
\begin{equation}
	TH_{JS}(T) = \frac{1}{n} \sum\limits_{d_i \in T} JS(P_{d_i},P_{T\setminus d_i})
\end{equation}
Here, $P_{d_i}$ is the frequency distribution of words in document $d_i$ and $P_{T \setminus d_i}$ is the frequency distribution of words in all other documents of the topic except $d_i$.
The final quantity $TH_{JS}$ is the average divergence of documents with all the others and provides, therefore, a measure of diversity among documents of a given topic.

\begin{table}[!tbp]
  \centering
  \begin{tabular}{lcccc}
  \toprule
  & BBC & Guardian & DUC '04 & TAC '08A\\
  \midrule
  $TH_{JS}$   & 0.5917 & 0.5689 & 0.3019 & 0.3188 	 \\
  \bottomrule
  \end{tabular}
  \caption{Average textual heterogeneity of our corpora compared to standard datasets}
  \label{tab:textual_heterogeneity}
\end{table}

We report the results in Table \ref{tab:textual_heterogeneity}.
To put the numbers in perspective, we also report the textual heterogeneity of 
the two standard summarization datasets DUC '04 and TAC '08A.
These corpora were created during shared tasks and focused on multi-document news summarization.
The heterogeneity in BBC and Guardian are similar and both much higher than DUC '04 and TAC '08A, meaning that our corpora contain more lexical variation inside topics. 

%% file: experiments.tex
\begin{table*}[!htbp]
\centering
\begin{tabular}{l*{4}{@{\hspace{.80cm}}c@{\hspace{.40cm}}c@{\hspace{.40cm}}c}}
\toprule
\multirow{2}{*}{Systems} & \multicolumn{3}{c}{BBC ($L$)}  & \multicolumn{3}{c}{Guardian ($L$)} & \multicolumn{3}{c}{BBC ($2*L$)} & \multicolumn{3}{c}{Guardian ($2*L$)} \\
 & R1 & R2 & SU4 & R1 & R2 & SU4 & R1 & R2 & SU4 & R1 & R2 & SU4\\
\midrule
TF*IDF    & .227 & .067 & .064 & .153 & .021 & .027 & .367 & .115 & .147 & .248 & .037 & .065 \\
LexRank   & .276 & .080 & .079 & .188 & .029 & .038 & .421 & .138 & .176 & .297 & .051 & .089\\
LSA       & .212 & .046 & .052 & .135 & .013 & .021 & .341 & .084 & .123 & .220 & .024 & .051 \\
KL        & .267 & .086 & .080 & .178 & .026 & .035 & .397 & .132 & .165 & .272 & .041 & .076 \\
ICSI      & .302 & .104 & .091 & .210 & .046 & .046 & .461 & .176 & .201 & .322 & .071 & .101\\
\midrule
UB-1      & \textbf{.514} & .273 & .218 & \textbf{.422} & .177 & .145 & \textbf{.754} & .388 & .435 & \textbf{.640} & .256 & .304\\
UB-2      & .494 & \textbf{.312} & .210 & .389 & \textbf{.230} & .137 & .709 & \textbf{.453} & .419 & .584 & \textbf{.334} & .277\\
\bottomrule
\end{tabular}
\caption{ROUGE-1 (R1), ROUGE-2 (R2), and ROUGE-SU4 (SU4) scores of multiple systems compared to the extractive upper bounds for ROUGE-1 (UB-1) and ROUGE-2 (UB-2) extractive for summary lengths of $L$ and $2*L$}
\label{tab:sumy:rouge}
\end{table*}

\begin{figure*}[!htbp]
\includegraphics[width=\textwidth]{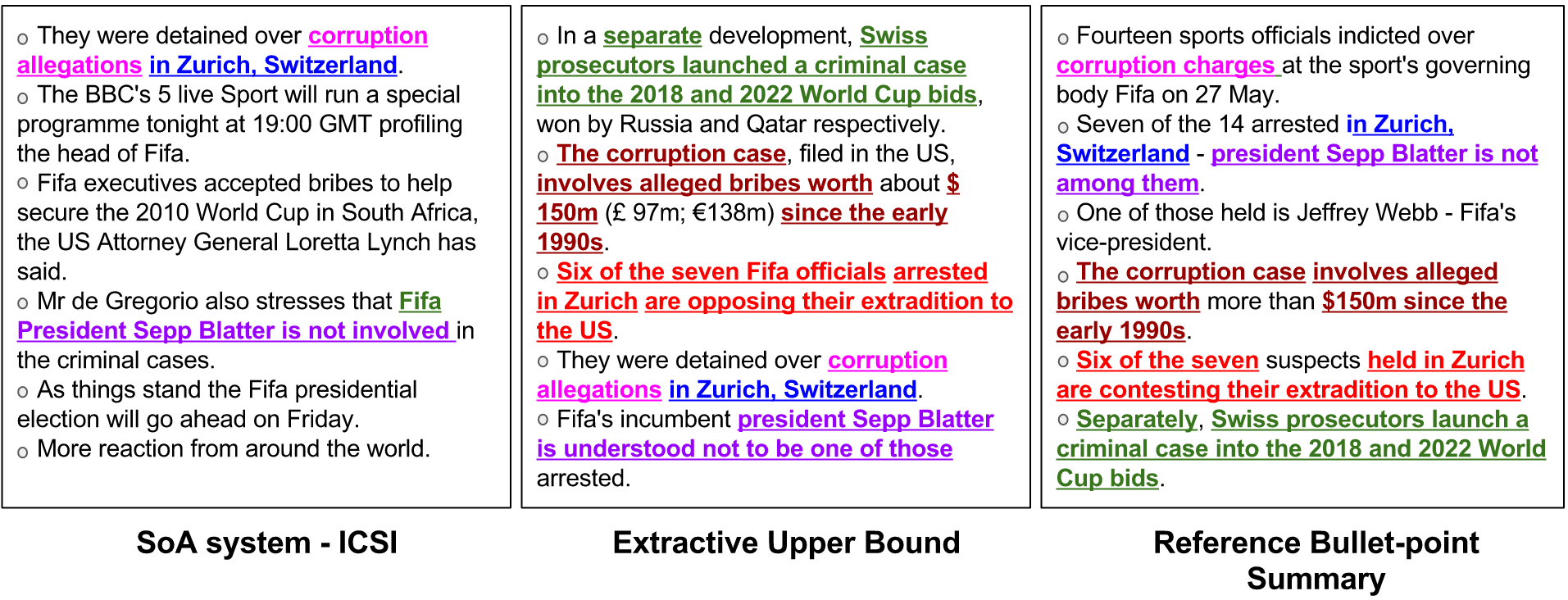}
\caption{BBC.com live blog on ``FIFA corruption inquiry"}
\label{fig:analysis}
\end{figure*}

In this section, we describe the automatic summarization methods and the upper bounds we compute for our live blog summarization dataset.

\subsection{Baselines} 
As benchmark results, we employ methods that have been successfully used for both single and multi-document summarization. 
Some variants of them have also been applied to update summarization tasks.

\textbf{TF$\star$IDF} \cite{Luhn} scores sentences with the TF*IDF of their terms. The best sentences are then greedily extracted.

\textbf{LexRank} \cite{Erkan2004} is a well-known graph-based approach.
A similarity graph $G(V,E)$ is constructed where $V$ is the set of sentences and an edge $e_{ij}$ is drawn between sentences $v_i$ and $v_j$ if and only if the cosine similarity between them is above a given threshold.
Sentences are then scored according to their PageRank in $G$. 

\textbf{LSA} \cite{steinberger2004using} is an approach involving a dimensionality reduction of the term-document matrix via singular value decomposition (SVD).
The sentences extracted should cover the most important latent topics.

\textbf{KL-Greedy} \cite{Haghighi:2009} minimizes the Kullback-Leibler (KL) divergence between the word distributions in summary and the documents. 

\textbf{ICSI} \cite{Gillick2009} is a global linear optimization that extracts a summary by solving a maximum coverage problem considering the most frequent bigrams in the source documents.
ICSI has been among the state-of-the-art MDS systems when evaluated with ROUGE \cite{HONG14}.

\subsection{Upper bound}
For comparison, we compute two upper bounds. 
The upper bound for extractive summarization is retrieved by solving the maximum coverage of n-grams from the reference summary \cite{TakamuraO10,TUD-ACL16,TUD-CS-2017-0077}.
This is cast as an Integer Linear Programming (ILP) and depends on two parameters: $N$, the size of n-grams considered and $L$, the maximum length of the summaries.
In our work, we set \(N=1\) and \(N=2\) and compute the upper bound for ROUGE-1 (UB-1) and ROUGE-2 (UB-2) respectively. 

\subsection{Experimental Setup}
We report scores for the ROUGE metrics identified by \newcite{Owczarzak:2012} as strongly correlating with human evaluation methods: ROUGE-1 (R1) and ROUGE-2 (R2) recall with stemming and stop words not removed.
For completeness, we also report the best skip-grams matching metric: ROUGE-SU4 (SU4). 

\subsection{Analysis}
Table \ref{tab:sumy:rouge} shows the results of benchmark summarization methods widely used in the summarization community on our live blog corpus. We explore two different summary lengths: 
$L$, length of the human-written bullet-point summary, and $2*L$, twice the length of the human-written summary to give leeway for compensating the excessive compression ratio of the human live blog summaries.
The results show the state-of-the-art ICSI system is .2 ROUGE-1 and .3 ROUGE-2 lower than the upper bounds for both BBC and the Guardian with length constraint $L$ and $2*L$ respectively.
ICSI is only able to reach one-third of the upper bound, which emphasizes that live blog summarization is a challenging task and we need new techniques tackling live blog summarization.

Figure \ref{fig:analysis} shows the output of the ICSI system as compared to the extractive upper bound on BBC live blog on ``FIFA corruption inquiry''.\footnote{\url{http://www.bbc.com/news/live/world-europe-32897157}}
It can be seen that the ICSI system extracts sentences with most frequent concepts (e.g., FIFA, president, world cup), but misses to identify topic shifts in these information snippets.
Although the information snippets collected by the ICSI system are related to FIFA corruption, it misses capturing relative importance of the information snippets.

Additionally, factors which determine the difficulty of the summarization task are the length of the source documents and the summary \cite{DBLP:conf/acl/NenkovaL08}.
The input document sizes of the BBC and the Guardian are on an average 5,890 and 6,048 words, whereas the summary sizes are around 59 and 42 words respectively.
Thus, the high compression ratio makes live blog summarization even more challenging.

%% file: conclusion.tex
We introduce a new task: live blog summarization which has direct applications for journalists and news readers.
Our goal is constructing a reference corpus for this new task.
In this paper, we suggest a pipeline to collect live blogs with human written bullet-point summaries from two major online newspapers, which can be extended to live blogs from other news agencies like \emph{The New York Times}, \emph{Washington Post} or \emph{Der Spiegel}.

We further analyze the live blog corpus and provide benchmark results for this dataset by applying commonly used summarization methods.
Our results show that off-the-shelf summarization systems cannot be used, as they are far from reaching the upper bound. 
This calls for new solutions that take the task characteristics into account.
As future work, we plan to research novel approaches to live blog summarization and investigate algorithms to identify important information from multiple topic shifts and a large number of information snippets.

Code for constructing and reproducing the live blog corpus and the automatic summarization experiments are published under the permissive Apache License 2.0 and can be obtained from \url{https://github.com/UKPLab/lrec2018-live-blog-corpus}.